\documentclass{article}
\usepackage{spconf,amsmath,epsfig}
\usepackage{amssymb}
\usepackage{amsthm}
\usepackage{color}
\usepackage{graphicx}

\begin{document}\sloppy

\def\x{{\mathbf x}}
\def\L{{\cal L}}

\title{Pedestrian re-identification based on Tree branch network with local and global learning}
%
\name{{Hui Li}$^1$, Meng Yang$^{2,*}$, Zhihui Lai$^1$, Weishi Zheng$^2$, Zitong Yu$^3$ \vspace{-1em}}
\address{$^1$Shenzhen University, $^2$Sun Yat-sen University, $^3$University of Oulu}

\maketitle

\begin{abstract}
Deep part-based methods in recent literature have revealed the great potential of learning local part-level representation for pedestrian image in the task of person re-identification. However, global features that capture discriminative holistic information of human body are usually ignored or not well exploited. This motivates us to investigate joint learning global and local features from pedestrian images.
Specifically, in this work, we propose a novel framework termed tree branch network (TBN) for person re-identification. Given a pedestrain image, the feature maps generated by the backbone CNN, are partitioned recursively into serveral pieces, each of which is followed by a bottleneck structure that learns finer-grained features for each level in the hierarchical tree-like framework. In this way, represenations are learned in a coarse-to-fine manner and finally assembled to produce more discriminative image descriptions.
Experimental results demonstrate the effectiveness of the global and local feature learning method in the proposed TBN framework. We also show significant improvement in performance over state-of-the-art methods on three public benchmarks: Market-1501, CUHK-03 and DukeMTMC.
\end{abstract}
\begin{keywords}
Pedestrian Re-Identification,  Tree Branch Network
\end{keywords}
\section{Introduction}
\label{sec:intro}

Pedestrian re-identification is a challenging task of computer vision to determine the presence of specific pedestrians in images or video sequences. Pedestrian re-identification faces many challenges \cite{wei2017glad,chen2016deep,wang2018learning}. For instance, the detection of pedestrian is easily affected by the accuracy of artificial markers or detection algorithms, the image appearance and resolution of pedestrian change dramatically as the lighting of the environment, the camera viewpoint, and the distance of the cameras to the target vary. Meanwhile pedestrians with the same identity can have quite in different appearance when captured different camera views. It is more difficult to distinguish their identities, when color of the clothes and the body shape among distinct pedestrians look similar.

Pedestrian re-identification has been studied by many researchers. In particular, many  part-based methods \cite{wei2017glad,yao2017deep,zhao2017deeply,sunbeyond,zhao2017spindle,li2017learning} have been verified as beneficial to person re-identification. They can be divided into two categories according to whether additional information (pose estimation, key point) needs to be introduced. Wei \emph{et al.} \cite{wei2017glad} explicitly leveraged the local and global cues in human body to generate a discriminative and robust representation with a few key points. Zhao \emph{et al.} \cite{zhao2017spindle} proposed a method based on human body region guided multi-stage feature decomposition, and tree-structured competitive feature fusion with more key points. These methods are overly dependent on the detection of key points or the estimation of poses. Adding extra information will inevitably lead to errors in detection and estimation. Yao \emph{et al.} \cite{yao2017deep} proposed part loss to learn more dicriminative local feature, which automatically detects human body part and computes each part separately. Zhao \emph{et al.} \cite{zhao2017deeply} proposed part-aligned human representation by detecting the human body regions and computing between the corresponding parts. Sun \emph{et al.} \cite{sunbeyond} proposed a strong convolution baseline method that leverages a uniform partition strategy. Although these methods do not require additional information, global discriminative features are not well explored together with local feature learning.

\begin{figure}[t]
\centering
\includegraphics[width=0.4\textwidth]{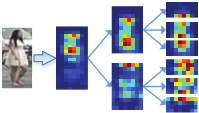}
\caption{Illustration of motivation of this paper. Global features are always not well exploited together with local features in existing deep part-based models. We argue that both of global and local features are critical, and should be jointly learned for person re-identification. The proposed framework, formulated as a hierarchical tree-branch structure, learns feature representations from different granularities and finally aggregates them to boost the performance.}
\label{fig:feature_visual}
\end{figure}
To address  the above drawbacks to some extent, in this paper, we propose a novel framework tree branch network (TBN) with local and global feature learning. To handle the variation of pedestrian, a tree branch network is designed to extract a discriminative feature including local and global information, without additional supervision information. What's more, we propose a feature learning strategy with local and global learning, for a powerful classifier of pedestrian images. Extensive experiments on Market1501, CUHK03 and DukeMTMC datasets are conducted, and clearly show the advantage of the proposed TBN in performance.

The contribution of this paper is as follows.
\begin{itemize}
    \item We propose a novel framework termed TBN with local and global feature learning. It does not require additional supervision information in the training procedure.
    \item We conduct extensive experiments to verify the effectiveness of tree branch network and demonstrate the superiority of our method.
\end{itemize}

\section{Related Work }
Feature learning and model learning are two important issues in pedestrian re-identification. In this section, we  review how to effectively use local or global feature, and how to learn a discriminative model in the literature of pedestrian re-identification.

\subsection{Local or Global Feature Extraction}

In \cite{sunbeyond,wei2017glad,zhao2017spindle}, the images of pedestrians are divided horizontally into different local regions, in which the body features of color, texture feature, and shape context are extracted. However, their robustness and discriminative power are still weak for the practical lighting and attitude changes.  As the emergence of deep learning, Zhao \emph{et al.} \cite{zhao2017spindle} proposed that images can be aligned macroscopically and microscopically by capturing semantic features from different body regions. In addition, the learned region features from different semantic regions are merged with a competitive scheme. Sun \emph{et al.}\cite{sunbeyond} proposed a unified segmentation strategy and conducted independent convolution with a refine part pooling. However, the network structure becomes relatively complex and will be damaged by the cumulative error of the key point information of the human body. In addition, \cite{sunbeyond} only considered the alignment of local information, ignoring the role of global information. Compared with directly partition the methods of each part, method of gradually partition can learn more discriminating features. Zhang \emph{et al.} \cite{zhang2017alignedreid} proposed to use the dynamic programming to find the shortest distance of the pedestrian image. Similarity, they also consider global information of pedestrian image and achieve promising accuracy with only global features. 

\subsection{Model Learning}
Deep learning \cite{yao2017deep,zhang2017alignedreid,zhao2017deeply,wang2018learning} is a main stream for pedestrian re-identification. Most approaches pay more attention to learn and aggregate local features \cite{zhao2017deeply,zhang2017alignedreid} achieve better performance by local alignment of the pedestrian image, and \cite{yao2017deep} design the loss of the local feature to obtain more discriminative features. The methed of \cite{wang2018learning} is to learn from coarse-grained to fine-grained features using multi-scale information.

Among these methods, mutual learning has achieved promising performance. The deep mutual learning proposed by Zhang \emph{et al.} \cite{zhang2018deep} is a typical model distillation method, which makes two deep networks learn from each other with Kullback-Leibler Divergence. {{As a very effective model distillation method, it achieves the goal of improving the effectiveness of a single student network by calculating the relative entropy between the different student models as mutual losses. 
}}The cooperative learning between two models will prefer to learn better generalization. Following \cite{zhang2018deep}, Zhang \emph{et al.} \cite{zhang2017alignedreid} adopted a mutual learning strategy to greatly improve the accuracy of pedestrian re-identification. Besides, Tong \emph{et al.} \cite{xiao2017joint} proposed to combined detection with identification learning, and they get a good performance with ID-discriminative embedding.

\section{Tree branch network learning with local and global feature learning}

To effectively use the joint information of local and global features and exploit the discrimination in the model learning, we propose a novel framework called Tree Branch Network (TBN) with local and global feature learning.  
In this section, we  first give an overview of TBN, then present the structure of tree branch network, followed by the decription of the optimization strategy in details.

\begin{figure}[t]
\centering
\includegraphics[width=0.45\textwidth]{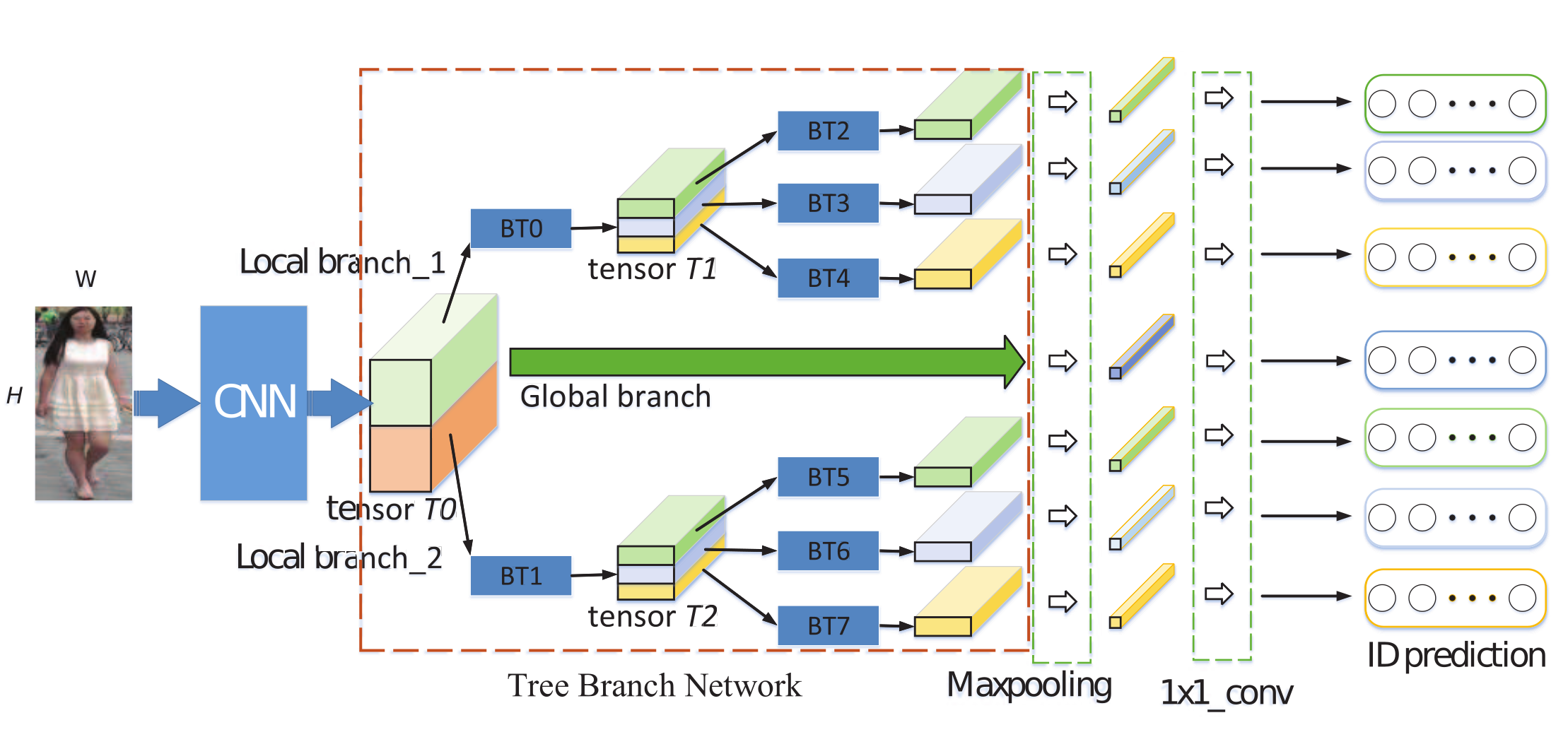}
\caption{Overview of the proposed TBN framework.}
\label{fig:1}
\end{figure}

\subsection{Overview of Proposed Framework }

Inspired by the idea of ``Divide and Conquer", we propose to learn a deep network from coarse to fine, and finally integrate all the processes in an objective loss. As shown in Fig.1, the given image is fed to the backbone network that outputs a 3D tensor. The proposed framework consists of two main modules: tree branch network and local and global learning. Through the processing of our proposed tree branch network, global feature in the global branch and multiple scale local features in local branches are extracted. After the pooling and 1$\times$1 kernel-sized convolutional layer, each part is trained for ID prediction with a softmax loss separately following \cite{sunbeyond}. Our proposed local and global learning module trains several powerful classifiers for the local and global features.
The proposed TBN module, including multiple branches to extract global and local information, provides a reasonable way of dividing the image feature tensor; while the objective for local and global learning can make full use of the joint probability distribution of all local and global features.

\subsection{Tree Branch Network (TBN) }
The TBN module, which learns an object from coarse to fine. So local information can be effectively extracted as the increase of partition.

A given image in each batch is fed to the backbone network that outputs a 3D tensor. TBN uniformly divides the original image tensor into two coarse tensors, which are further divided into six tensors. Here the process of division also includes the network of bottleneck. Bottleneck maintains the size of the image when extracting local information because it conducts convolution and keeps spatial information at the same time. Compared to the partition which directly divides the image into six local tensors, larger reception fields are preserved and different regions are given an adaptive attention in the feature extraction. {{Compared to PCB \cite{sunbeyond}, our advantage is that our blocking method is a coarse-to-fine process. This allows us to learn better local features. Meanwhile, our blocking approach is not simple feature space segmentation. The bottleneck structure allows us to integrate the features of each tile and strengthen the connection between the channels. 
}}

\begin{figure}[t]
\centering

\includegraphics[width=0.4\textwidth]{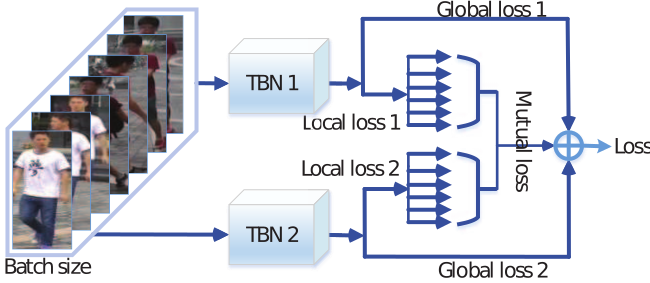}
\caption{Illustration of mutual learning. 
Each batch of images are fed to two modules of our proposed TBN simultaneously.  Based on the two networks, e.g., TBN$_{1}$ and TBN$_{2}$, the corresponding local and global learning loss can be calculated, respectively. 
}
\label{fig:3}
\end{figure}  

\begin{table*}[t]
\begin{center}
\caption{ Experiment results of our method. ``+" stands for experiments with classification mutual loss. The dim stands for the dimension of the feature. ``local" indicates that all local branch features are concated, and ``global" indicates the feature of the global branch.} \label{tab:OBF}
\small
\begin{tabular}{c|c|c|c|c|c|c|c}
  \hline
    Method  & dim  &   \multicolumn{2}{c|}{Market501} &  \multicolumn{2}{c|}{CUHK03}  &  \multicolumn{2}{c}{DukeMTMC-reID}  \\
     &  & \multicolumn{1}{c}{Rank-1} & mAP & \multicolumn{1}{c}{Rank-1} & mAP &  \multicolumn{1}{c}{Rank-1} & mAP  \\\hline\hline
    TBN (local only) & 1536 &\multicolumn{1}{c}{92.7}& 78.9 &\multicolumn{1}{c}{55.2}& 50.5 & \multicolumn{1}{c}{85.0} &  70.7 \\
    TBN (global only) & 2048 &\multicolumn{1}{c}{91.7}& 78.5 &\multicolumn{1}{c}{56.2}& 51.2 & \multicolumn{1}{c}{82.6} &  66.6 
    \\
    TBN & 3584 &\multicolumn{1}{c}{92.9}& 79.9 &\multicolumn{1}{c}{59.1}& 54.5 & \multicolumn{1}{c}{85.2} &  71.5 \\
    TBN+ & 3584 &\multicolumn{1}{c}{93.2}& 83.0 &\multicolumn{1}{c}{69.2}& 65.0 & \multicolumn{1}{c}{85.5} &  73.0 \\
  \hline
\end{tabular}
\end{center}
\vspace{-2em}
\end{table*}

\subsection{Optimization}
For the network learning, we propose an objective for local and global learning, which can make full use of the joint probability distribution of all local and global features. As shown in Fig.(\ref{fig:1}), the loss function defined as 
\begin{eqnarray}
L = L_{local}+L_{global}
\end{eqnarray} 
The first term of Eq.(1) represents the sum of all local cross entropy losses 
\begin{eqnarray}
L_{local}=-\frac{1}{N}\sum_{i=0}^{N}\sum_{k=0}^{K}\sum_{m=0}^{M}I\left ( y_{i},m \right )h\left ( x_{i}^{k} \right )
\end{eqnarray}  
where $h\left ( x_{i}^{k} \right )=\log \left ( \exp{({f_{j}^{k})}} \right/\sum _{m=1}^{M}\exp{({f_{m}^{k}})} )$ , a label set is represented $\left \{ y_{i} \right \}_{i=1}^{N}$ as with $y\in \left \{ 1,2,...M \right \}$ , and an indicator function is defined as $I(y_{i},m) =\left\{\begin{array}{lcl}1\quad & y = m \\0\quad &y\neq m\end{array}\right.$ 
The second term of Eq.(1),  represents the cross entropy loss for the global branch. We defined  as global loss
\begin{eqnarray}
L_{global}=-\frac{1}{N}\sum_{i=0}^{N}\sum_{m=0}^{M}I\left ( y_{i},m \right )g\left ( x_{i} \right )
\end{eqnarray}
where $ g\left ( x_{i} \right )=\log\left ( \exp{({d_{j}})} \right/\sum _{m=1}^{M}\exp{({d_{m}})} )$ , and the logit $d_{j}$ is input of $i$-th sample to the softmax layer in the global branch TBN.

\noindent\textbf{Mutual learning}. Inspired by the idea in \cite{zhang2018deep}, we build two identical models to more effectively learn image features. As shown in Fig.(\ref{fig:3}), there are two modules in our propose TBN, in which mutual loss can be applied to jointly learning the networks of TBN$_{1}$ and TBN$_{2}$. Take the network of TBN$_{1}$ as an example.

\noindent The overall loss function is defined as
\begin{eqnarray}
L_{_{\Theta _{1}}}= L_{kl}+L_{1}
\end{eqnarray}   
where $L_{kl}$ represents the local mutual loss of two TBN models, $L_{1}$ represents the loss of  of single TBN$_{1}$. Here $L_{kl}$ is defined as
\begin{eqnarray}
L_{kl}=D_{KL}\left ( p||q \right )=\frac{1}{N}\sum_{i=1}^{N}p\left ( x_{i} \right )\log\frac{p\left ( x_{i} \right )}{q\left ( x_{i} \right )}
\end{eqnarray}
where $p\left ( x_{i} \right )$ and $q\left ( x_{i} \right )$ are the output possibilities of TBN$_{1}$ and TBN$_{2}$, respectively. For instance, 
\begin{eqnarray}
p\left ( x_{i} \right )=\frac{\exp{({F_{i}})}}{\sum _{m=1}^{M}\exp{({F_{m}})}}
\end{eqnarray}
where $F_{i}=\left [ f_{i}^{1},\cdots,f_{i}^{k},\cdots,f_{i}^{K} \right ]$
is the concatenation of all logits of local branch probabilities, and the logit $f_{i}^{k}$ is input of i-th sample to the softmax layer in the k-th local branch TBN$_{1}$, and K represents the final partition number of part.
Compared with the logit of each local branch to calculate a mutual loss, the logit concate of all local branches together to calculate
the mutual loss can reduce the over-fitting. The effect of concatenation of all the local feature can make the objective smaller. 

Similarly, the objective loss function for network TBN$_{2}$ can be computed as
\begin{eqnarray}
L_{_{\Theta _{2}}}= L_{kl}+L_{2}
\end{eqnarray}     

\section{Experiment}

In this section, we evaluate the proposed approach for the task of pedestrian re-identification on three benchmark datasets. Extensive comparisons are conducted and compared with methods include many state-of-the-art methods, such as GLAD \cite{wei2017glad}, MSCAN \cite{li2017learning}, DLPA\cite{zhao2017deeply}, SVDNet \cite{sun2017svdnet}, PDC \cite{su2017pose}, TriNet \cite{hermans2017defense}, JLML \cite{li2017person}, DML \cite{zhang2018deep}, DPFL\cite{chen2017person}, HA-CNN \cite{li2018harmonious}, GP-reid \cite{almazan2018re}, PCB\cite{sunbeyond}, Deep-Person \cite{bai2017deep}, PCB+RPP \cite{sunbeyond}, and AlignedReID \cite{zhang2017alignedreid}.
\subsection{Datasets and Protocols}

\textbf{Market-1501} dataset \cite{zheng2015scalable}  comes from 6 different cameras and contains 1,501 different identities and 19,732 gallery images. The training set contains 12,936 images that we use for training. All images are detected by DPM \cite{felzenszwalb2008discriminatively}.

\noindent\textbf{CUHK03} \cite{li2014deepreid} has two types of bounding boxes which are hand-labeled and DPM-detected. We use both ways to validation our method. This dataset is randomly split 20 training/test. We adopt new training/test protocol \cite{zhong2017re} to validate our method.

\noindent\textbf{DukeMTMC-reID}  \cite{zheng2018pedestrian} is a pedestrian re-identification subset of the DukeMTMC dataset collected from 8 different cameras and provides a manually labeled bounding box. It contains 16,522 training images including 702 people, 2,228 query images from another 702 people, and 17,661 images as gallery.

\subsection{Implementation Details}
We use ResNet50 ImageNet \cite{deng2009imagenet} pre-trained model and remove layers after global pooling layer. We also change the step size of the last convolutional layer. So size of tensor T$_{0}$ is 24$\times$8, and the size of tensor$_{1}$ or tensor T$_{2}$ is 12$\times$8. The size of the remaining tensors before maxpooling is 4$\times$8 To learn more discriminative features, we use bottleneck structures to learn the local features of each part of the image. The bottleneck structure is a legacy residual network. It consists of a 1$\times$1, 3$\times$3, and 1$\times$1 convolution with a ReLU function between each convolutional layer. 1$\times$1 convolution is used to change the dimensions of the feature, and a 3$\times$3 convolution is used to increase the receptive field of the local features of the image.

Following \cite{sunbeyond}, all images are resized to 384$\times$128. Batch size is set to 64. As shown in Fig.(\ref{fig:1}), we train the model with a base learning rate of 0.01 for pre-trained model layers. The rest of layers are trained with a learning rate of 0.1. The learning rates are divided by 10 after 40 epoch. The number of epochs for training is set as 60. For mutual learning, we train model for 300 epochs with base learning rates of 0.02 and 0.002 for pre-trained model layers and the rest layers, respectively. Learning rates are divided by 10 after the 150$_{th}$ epoch until the convergence of learning. Experiments are implemented on pytorch platform with two NVIDIA TITAN XP GPUs. 



\subsection{Performance Evaluation}

Table 1 presents the comparison results of several variants of the proposed TBN framework on all of three benchmark datasets. We have the following observations. (1) Learning and assembling local feature representations are beneficial to better performance. This has been verified by existing deep part-based models. Note that, on Market-1501 and Duke-MTMC datasets, the model employing only local features performs even better than its variant using only global features. The possible reason is that local features contain more spatial-related discriminative information that leads to performance gain. (2) The model that learns joint global and local features performs better than using either of them, on all of three datasets. It demonstrates the effectiveness of using both global and local features learned in a coarse-to-fine manner, because feature representation learned from different granularities are complementary. It is for this reason, we formulate learning global and local features in the proposed TBN, a hierarchal tree-like structure. (3) Mutual learning also boosts the performance, especially the improvement on the CUHK-03 dataset, which verifies its great importance in our proposed framework.


\subsection{Comparison with State-of-the-Art Methods}

We present the comparison results on the Market-1501 dataset in Table 2. Our proposed TBN achieves best performance among all the state-of-the-art methods, except PCB+RPP \cite{sunbeyond} in the case of single query. PCB+RPP is a very strong competitor. Compared to PCB+RPP, the TBN has a similar Rank-1 accuracy but much better mAP accuracy (1.4\% improvement in mAP, which reflects the ability to recall all samples and can better reflect the superiority of the method). That shows with the same backbone (Resnet50), our partition method is more effective. When with the re-ranking operation \cite{zhong2017re}, the proposed TBN(RK) has achieved the state-of-the-art performance, with at least 1\% improvement in all cases. Figure 4 shows the top-10 ranking for some query images on Market-1501 dataset by TBN.
\begin{table}[t]
\small
\begin{center}
\caption{Comparison of results on Market-1501 with evaluation protocols in \cite{zhong2017re}. ``RK" stands for re-ranking operation.}
\begin{tabular}{c|c|c|c|c}
  \hline
    Method  &  \multicolumn{2}{c|}{Single Query} &  \multicolumn{2}{c}{Multiple Query} \\
     &  \multicolumn{1}{c}{Rank-1} & mAP & \multicolumn{1}{c}{Rank-1} & mAP   \\\hline\hline
   GLAD \cite{wei2017glad} & \multicolumn{1}{c}{73.9}& 89.9 &\multicolumn{1}{c}{-}& -\\
   MSCAN \cite{li2017learning}  & \multicolumn{1}{c}{80.3}& 57.5 &\multicolumn{1}{c}{86.8}& 66.7\\
   DLPA \cite{zhao2017deeply}  & \multicolumn{1}{c}{81.3}& 63.4 &\multicolumn{1}{c}{-}& -\\
   SVDNet \cite{sun2017svdnet}  & \multicolumn{1}{c}{82.3}& 62.1 &\multicolumn{1}{c}{-}& -\\
   PDC \cite{su2017pose}  & \multicolumn{1}{c}{84.1}& 69.1 &\multicolumn{1}{c}{-}& -\\
   TriNet \cite{hermans2017defense}   & \multicolumn{1}{c}{84.9}& 69.1 &\multicolumn{1}{c}{90.5}& 76.4\\
   JLML \cite{li2017person}  & \multicolumn{1}{c}{85.1}& 65.5 &\multicolumn{1}{c}{89.7}& 74.5\\
   DML \cite{zhang2018deep}   & \multicolumn{1}{c}{87.7}& 68.8 &\multicolumn{1}{c}{91.7}& 77.7\\
   DPFL \cite{chen2017person}  & \multicolumn{1}{c}{88.6}& 72.6 &\multicolumn{1}{c}{92.2}& 80.4\\
   HA-CNN \cite{li2018harmonious}  & \multicolumn{1}{c}{91.2}& 75.7 &\multicolumn{1}{c}{93.8}& 82.8\\
   GP-reid \cite{almazan2018re}  & \multicolumn{1}{c}{92.2}& 81.2 &\multicolumn{1}{c}{94.7}& 87.3\\
   Deep-Person \cite{bai2017deep}  & \multicolumn{1}{c}{92.3}& 79.6 &\multicolumn{1}{c}{94.5}& 85.1\\
   PCB+RPP \cite{sunbeyond}  & \multicolumn{1}{c}{\textbf{93.8}}& 81.6 &\multicolumn{1}{c}{-}& -\\
   \textbf{TBN+}  & \multicolumn{1}{c}{93.2}& \textbf{83.0} &\multicolumn{1}{c}{\textbf{95.4}}& \textbf{88.0}\\ 
   \hline\hline
   GP-reid (RK) \cite{almazan2018re}   & \multicolumn{1}{c}{92.2}& 90.0 &\multicolumn{1}{c}{94.2}& 91.2\\ 
   AlignedReID (RK) \cite{zhang2017alignedreid} & \multicolumn{1}{c}{94.4}& 90.7 &\multicolumn{1}{c}{-}& -\\ 
   \textbf{TBN+} (RK)  & \multicolumn{1}{c}{\textbf{95.4}}& \textbf{91.3} &\multicolumn{1}{c}{\textbf{95.8}}& \textbf{94.0}\\ 
  \hline
\end{tabular}
\end{center}
\vspace{-2em}
\end{table}

CUHK03 is a very challenging dataset, on which the comparison has been conducted with the performance listed in Table 3.  There are two methods of pedestrian boxes. We adopt the new training/testing protocol proposed in \cite{zhong2017re}. Compared to PCB+RPP, our method improves 5.5\% and 7.5\% without re-ranking. Compared to MGN \cite{wang2018learning}, our method exceeds 2.4\% in the accuracy of rank-1. Our method achieves Rank-1/mAP = 72.3\%/68.5\% on CUHK03 labeled setting which outperform all the published results by a large margin.
\begin{table}[t]
\small
\begin{center}
\caption{ Comparison of results on CUHK03 with evaluation protocols in [46] in single query mode. ``RK" stands for re-ranking operation. }
\begin{tabular}{c|c|c|c|c}
  \hline
    Method  &  \multicolumn{2}{c|}{Labeled} &  \multicolumn{2}{c}{Detected} \\
     &  \multicolumn{1}{c}{Rank-1} & mAP & \multicolumn{1}{c}{Rank-1} & mAP   \\\hline\hline
   IDE \cite{zheng2016person} & \multicolumn{1}{c}{22.2}& 21.0 &\multicolumn{1}{c}{21.3}& 19.7\\
   PAN \cite{zheng2018pedestrian}  & \multicolumn{1}{c}{36.9}& 35.0 &\multicolumn{1}{c}{36.3}& 34\\
   MultiScale \cite{chen2017person}  & \multicolumn{1}{c}{-}& - &\multicolumn{1}{c}{40.7}& 37.0\\
   SVDNet \cite{sun2017svdnet}  & \multicolumn{1}{c}{40.9}& 37.8 &\multicolumn{1}{c}{41.5}& 37.3\\
   HA-CNN \cite{li2018harmonious} & \multicolumn{1}{c}{44.4}& 41.0 &\multicolumn{1}{c}{41.7}& 38.6 \\
   SVDNet+Era \cite{zhong2017random}  & \multicolumn{1}{c}{-}& - &\multicolumn{1}{c}{48.7}& 43.5\\
   MLFN \cite{chang2018multi} & \multicolumn{1}{c}{54.7}& 49.2 &\multicolumn{1}{c}{52.8}& 47.8\\
   TriNet+Era \cite{zhong2017random} & \multicolumn{1}{c}{-}& - &\multicolumn{1}{c}{55.5}& 50.7\\
   PCB+RPP \cite{sunbeyond} & \multicolumn{1}{c}{-}& - &\multicolumn{1}{c}{63.7}& 57.5\\ 
   MGN \cite{wang2018learning} & \multicolumn{1}{c}{68.0}& 67.4 &\multicolumn{1}{c}{66.8}& \textbf{66.0}\\ 
   \hline\hline
   \textbf{TBN+}  & \multicolumn{1}{c}{\textbf{72.3}}& \textbf{68.5} &\multicolumn{1}{c}{\textbf{69.2}}& 65.0\\ 
   \textbf{TBN+ (RK)}  & \multicolumn{1}{c}{\textbf{79.5}}& \textbf{80.7} &\multicolumn{1}{c}{\textbf{75.9}}& \textbf{77.5}\\ 
  \hline
\end{tabular}
\end{center}
\end{table}

We evaluate all the competing methods on the challenging DukeMTMC-reID dataset, and the comparison results are listed in Table 4. From Table 4, it can be observed that our method performs excellently again. Compared to some existing methods with multi-scale methods, the performance achieved by our TBN still surpasses them significantly. Our TBN achieves Rank-1/mAP =73.0\%/85.5\%, outperforming SphereReID \cite{fan2019spherereid} by +1.6\% in Rank-1 and +4.5\% in mAP. Compared to PCB+RPP, our method exceeds 2.6\% in  rank-1 and 4.9\%. GP-reid \cite{almazan2018re} is a good method on Reid-task. It achieves quite good performance, but our method still show better performance than GP-reid. Similar to Table 2 and Table 3, with the re-ranking operation, the accuracy of the TBN is further enhanced, with the state-of-the-art performance reported.
\begin{table}[t]
\small
\begin{center}
 \caption{ Comparison of results on DukeMTMC-reID in single query mode.}
\begin{tabular}{c|c|c}
  \hline
    Method  &  \multicolumn{1}{c}{Rank-1} & mAP  \\\hline\hline
   GAN \cite{zheng2017unlabeled} & \multicolumn{1}{c}{67.7}& 47.1\\
   PAN \cite{zheng2018pedestrian} & \multicolumn{1}{c}{71.6}& 51.5\\
   SVDNet \cite{sun2017svdnet} & \multicolumn{1}{c}{76.7}& 56.8\\
   MultiSacle \cite{chen2017person}  & \multicolumn{1}{c}{79.2}& 60.6\\
   TriNet+Era \cite{zhong2017random}  & \multicolumn{1}{c}{73.0}& 56.6\\
   SVDNet+Era \cite{zhong2017random} & \multicolumn{1}{c}{79.3}& 62.4\\
   HA-CNN \cite{li2018harmonious} & \multicolumn{1}{c}{80.5}& 63.8\\
   PCB \cite{sunbeyond}  & \multicolumn{1}{c}{81.8}& 66.1\\
   PCB+RPP \cite{sunbeyond} & \multicolumn{1}{c}{82.9}& 68.1\\
   SphereReID \cite{fan2019spherereid}  & \multicolumn{1}{c}{83.9}& 68.5\\
   GP-reid \cite{almazan2018re}   & \multicolumn{1}{c}{85.2}& 72.8\\
   \hline\hline
   \textbf{TBN+}  & \multicolumn{1}{c}{\textbf{85.5}}& \textbf{73.0} \\ 
   \textbf{TBN+ (RK)}  & \multicolumn{1}{c}{\textbf{89.2}}& \textbf{85.1} \\ 
  \hline
\end{tabular}  
\end{center}
\end{table}

\begin{figure}[t]
\centering
\includegraphics[width=0.4\textwidth]{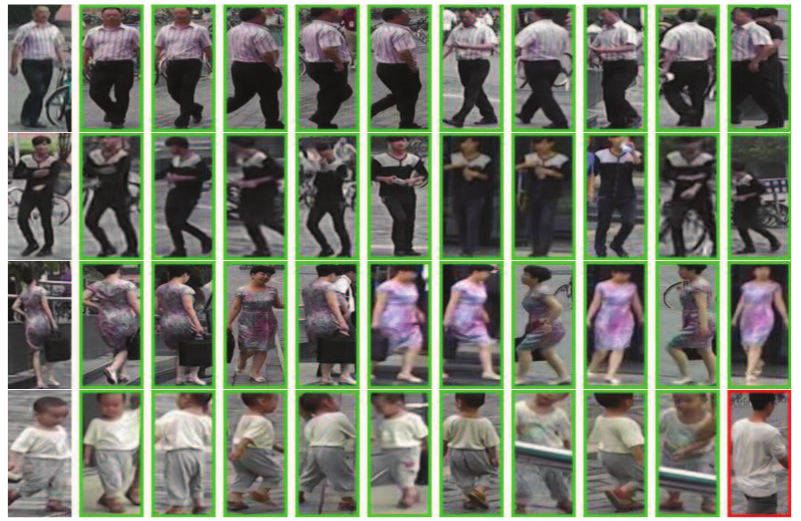}
\caption{The top-10 ranking list for some query images on Market-1501 dataset by our model. On the far left is the query image. Green boundary is added to true positive, and red to false positive.}
\label{fig:4}
\end{figure}

\section{CONCLUSION}

In this paper, we propose a method of Tree Branch Network (TBN) with local and global feature learning  for person re-identification tasks. Compared with the previous methods, our proposed method pays more attention to the learning of global and local features. A novel module of local and global learning is also presented, resulting in several powerful classifiers of pedestrian images. Extensive experiments on Market1501 dataset, CUHK03 and DukeMTMC clearly show that the proposed TBN has achieved state-of-the-art performance on several benchmark datasets.


\setlength{\baselineskip}{0.5em}    

\let\oldbibliography\thebibliography
\renewcommand{\thebibliography}[1]{%
  \oldbibliography{#1}%
  \setlength{\parskip}{0.5pt}
  \setlength{\itemsep}{-3pt}
}

\bibliographystyle{IEEEbib}
\small
\bibliography{icme2019template}

\begin{thebibliography}{10}

\bibitem{wei2017glad}
Longhui Wei, Shiliang Zhang, Hantao Yao, Wen Gao, and Qi~Tian,
\newblock ``Glad: global-local-alignment descriptor for pedestrian retrieval,''
\newblock in {\em ACM MM}, 2017.

\bibitem{chen2016deep}
Shi-Zhe Chen, Chun-Chao Guo, and Jian-Huang Lai,
\newblock ``Deep ranking for person re-identification via joint representation
  learning,''
\newblock {\em IEEE Transactions on Image Processing}, 2016.

\bibitem{wang2018learning}
Guanshuo Wang, Yufeng Yuan, Xiong Chen, Jiwei Li, and Xi~Zhou,
\newblock ``Learning discriminative features with multiple granularities for
  person re-identification,''
\newblock in {\em ACM MM}, 2018.

\bibitem{yao2017deep}
Hantao Yao, Shiliang Zhang, Yongdong Zhang, Jintao Li, and Qi~Tian,
\newblock ``Deep representation learning with part loss for person
  re-identification,''
\newblock {\em arXiv preprint}, 2017.

\bibitem{zhao2017deeply}
Liming Zhao, Xi~Li, Yueting Zhuang, and Jingdong Wang,
\newblock ``Deeply-learned part-aligned representations for person
  re-identification.,''
\newblock in {\em ICCV}, 2017.

\bibitem{sunbeyond}
Yifan Sun, Liang Zheng, Yi~Yang, Qi~Tian, and Shengjin Wang,
\newblock ``Beyond part models: Person retrieval with refined part pooling (and
  a strong convolutional baseline),''
\newblock in {\em ECCV}, 2017.

\bibitem{zhao2017spindle}
Haiyu Zhao, Maoqing Tian, Shuyang Sun, Jing Shao, Junjie Yan, Shuai Yi,
  Xiaogang Wang, and Xiaoou Tang,
\newblock ``Spindle net: Person re-identification with human body region guided
  feature decomposition and fusion,''
\newblock in {\em CVPR}, 2017.

\bibitem{li2017learning}
Dangwei Li, Xiaotang Chen, Zhang Zhang, and Kaiqi Huang,
\newblock ``Learning deep context-aware features over body and latent parts for
  person re-identification,''
\newblock in {\em CVPR}, 2017.

\bibitem{zhang2017alignedreid}
Xuan Zhang, Hao Luo, Xing Fan, Weilai Xiang, Yixiao Sun, Qiqi Xiao, Wei Jiang,
  Chi Zhang, and Jian Sun,
\newblock ``Alignedreid: Surpassing human-level performance in person
  re-identification,''
\newblock {\em arXiv preprint}, 2017.

\bibitem{zhang2018deep}
Ying Zhang, Tao Xiang, Timothy~M Hospedales, and Huchuan Lu,
\newblock ``Deep mutual learning,''
\newblock in {\em CVPR}, 2018.

\bibitem{xiao2017joint}
Tong Xiao, Shuang Li, Bochao Wang, Liang Lin, and Xiaogang Wang,
\newblock ``Joint detection and identification feature learning for person
  search,''
\newblock in {\em CVPR}, 2017.

\bibitem{sun2017svdnet}
Yifan Sun, Liang Zheng, Weijian Deng, and Shengjin Wang,
\newblock ``Svdnet for pedestrian retrieval,''
\newblock {\em ICCV}, 2017.

\bibitem{su2017pose}
Chi Su, Jianing Li, Shiliang Zhang, Junliang Xing, Wen Gao, and Qi~Tian,
\newblock ``Pose-driven deep convolutional model for person
  re-identification,''
\newblock in {\em ICCV}, 2017.

\bibitem{hermans2017defense}
Alexander Hermans, Lucas Beyer, and Bastian Leibe,
\newblock ``In defense of the triplet loss for person re-identification,''
\newblock {\em arXiv preprint}, 2017.

\bibitem{li2017person}
Wei Li, Xiatian Zhu, and Shaogang Gong,
\newblock ``Person re-identification by deep joint learning of multi-loss
  classification,''
\newblock {\em arXiv preprint}, 2017.

\bibitem{chen2017person}
Yanbei Chen, Xiatian Zhu, and Shaogang Gong,
\newblock ``Person re-identification by deep learning multi-scale
  representations,''
\newblock in {\em Proceedings of the IEEE International Conference on Computer
  Vision}, 2017, pp. 2590--2600.

\bibitem{li2018harmonious}
Wei Li, Xiatian Zhu, and Shaogang Gong,
\newblock ``Harmonious attention network for person re-identification,''
\newblock in {\em CVPR}, 2018.

\bibitem{almazan2018re}
Jon Almazan, Bojana Gajic, Naila Murray, and Diane Larlus,
\newblock ``Re-id done right: towards good practices for person
  re-identification,''
\newblock {\em arXiv preprint}, 2018.

\bibitem{bai2017deep}
Xiang Bai, Mingkun Yang, Tengteng Huang, Zhiyong Dou, Rui Yu, and Yongchao Xu,
\newblock ``Deep-person: Learning discriminative deep features for person
  re-identification,''
\newblock {\em arXiv preprint}, 2017.

\bibitem{zheng2015scalable}
Liang Zheng, Liyue Shen, Lu~Tian, Shengjin Wang, Jingdong Wang, and Qi~Tian,
\newblock ``Scalable person re-identification: A benchmark,''
\newblock in {\em ICCV}, 2015.

\bibitem{felzenszwalb2008discriminatively}
Pedro Felzenszwalb, David McAllester, and Deva Ramanan,
\newblock ``A discriminatively trained, multiscale, deformable part model,''
\newblock in {\em CVPR}, 2008.

\bibitem{li2014deepreid}
Wei Li, Rui Zhao, Tong Xiao, and Xiaogang Wang,
\newblock ``Deepreid: Deep filter pairing neural network for person
  re-identification,''
\newblock in {\em CVPR}, 2014.

\bibitem{zhong2017re}
Zhun Zhong, Liang Zheng, Donglin Cao, and Shaozi Li,
\newblock ``Re-ranking person re-identification with k-reciprocal encoding,''
\newblock in {\em CVPR}, 2017.

\bibitem{zheng2018pedestrian}
Zhedong Zheng, Liang Zheng, and Yi~Yang,
\newblock ``Pedestrian alignment network for large-scale person
  re-identification,''
\newblock {\em IEEE Transactions on Circuits and Systems for Video Technology},
  2018.

\bibitem{deng2009imagenet}
Jia Deng, Wei Dong, Richard Socher, Li-Jia Li, Kai Li, and Li~Fei-Fei,
\newblock ``Imagenet: A large-scale hierarchical image database,''
\newblock in {\em CVPR}, 2009.

\bibitem{zheng2016person}
Liang Zheng, Yi~Yang, and Alexander~G Hauptmann,
\newblock ``Person re-identification: Past, present and future,''
\newblock {\em arXiv preprint}, 2016.

\bibitem{zhong2017random}
Zhun Zhong, Liang Zheng, Guoliang Kang, Shaozi Li, and Yi~Yang,
\newblock ``Random erasing data augmentation,''
\newblock {\em arXiv preprint}, 2017.

\bibitem{chang2018multi}
Xiaobin Chang, Timothy~M Hospedales, and Tao Xiang,
\newblock ``Multi-level factorisation net for person re-identification,''
\newblock in {\em CVPR}, 2018.

\bibitem{fan2019spherereid}
Xing Fan, Wei Jiang, Hao Luo, and Mengjuan Fei,
\newblock ``Spherereid: Deep hypersphere manifold embedding for person
  re-identification,''
\newblock {\em Journal of Visual Communication and Image Representation}, 2019.

\bibitem{zheng2017unlabeled}
Zhedong Zheng, Liang Zheng, and Yi~Yang,
\newblock ``Unlabeled samples generated by gan improve the person
  re-identification baseline in vitro,''
\newblock in {\em ICCV}, 2017.

\end{thebibliography}
\end{document}